# The Novel Approach of Adaptive Twin Probability for Genetic Algorithm

Anagha P. Khedkar
Associate Professor, I.T. Department,
Matoshri College of Engineering and Research,
Nashik, India

Dr. Shaila Subbaraman
Ex-Academic Dean, Electronics Department
Walchand College of Engineering
Sangli, India

*Abstract*—The performance of GA is measured and analyzed in terms of its performance parameters against variations in its genetic operators and associated parameters. Since last four decades huge numbers of researchers have been working on the performance of GA and its enhancement. This earlier research work on analyzing the performance of GA enforces the need to further investigate the exploration and exploitation characteristics and observe its impact on the behavior and overall performance of GA. This paper introduces the novel approach of adaptive twin probability associated with the advanced twin operator that enhances the performance of GA. The design of the advanced twin operator is extrapolated from the twin offspring birth due to single ovulation in natural genetic systems as mentioned in the earlier works. The twin probability of this operator is adaptively varied based on the fitness of best individual thereby relieving the GA user from statically defining its value. This novel approach of adaptive twin probability is experimented and tested on the standard benchmark optimization test functions. The experimental results show the increased accuracy in terms of the best individual and reduced convergence time.

## I. INTRODUCTION

GAs are heuristic computational procedures based on the natural genetics and achieve a good compromise between deterministic approach and completely random probabilistic approach. The real strength of GA lies in its capabilities for solving complex hard optimization problems in all facets of real world. The major capability of GA is associated with its various operators and associated parameters. The primary genetic operators viz. selection and crossover play an important role in determining the behavior and performance of GA [1]. In addition to this, the secondary mutation operator significantly contributes to the exploration characteristics of GA.

Lot of research has been done in past on the various types of genetic operators and its associated parameters. The brief history of this is presented below. The selection operator and its various types viz. roulette wheel, tournament, stochastic apply different levels of selection pressure during the selection of individuals [2]. The selection pressure indirectly affects the convergence properties of GA. The crossover operator and its various types have been empirically studied and analyzed for observing their effects on the behavior of GA [3], [4], [5], [6]. In addition to different crossover types, the novel technique of crossover has been devised based on the variation in the number of parents and the number of crossovers [7]-[15]. The hybridization of various crossover operators has been tried for studying the synergetic effects [16], [17]. The probability of crossover $p_c$ also plays an important role in exploring the characteristics of GA. The numbers of attempts were executed for static, dynamic and adaptive approaches of crossover probability $p_c$ [18], [19], [20], [21]. On the parallel lines of crossover operator, many researchers have also empirically studied the secondary mutation operator and its different types. The probability of mutation $p_m$ along with the mutation technique affects the diversity of the population. The static, dynamic and adaptive approaches for $p_m$ have been attempted and analyzed in the GA literature [22], [23], [24]. The variations in crossover and mutation probabilities significantly affect the exploration and exploitation properties of GA.

Recently in 2011, Matej et al. carried out the thorough analysis of exploration and exploitation characteristics of GA and its results are presented in terms of basic understanding of these terms along with the performance measures [25]. They reviewed large number of research papers on evolutionary algorithms especially focusing on the exploration and exploitation characteristics. The effects of these characteristics based on the various criteria viz. stage of application, way of application are presented. This review highlights the future scope for further controlling and balancing exploration and exploitation characteristics of GA by incorporating the new directions. This remark in combination with the various adaptive approaches for crossover and mutation probabilities has motivated to propose the novel approach of adaptive twin probability for the advanced twin operator of GA.

Since last four decades GA has been under continuous development phase especially in terms of its operators and parameters. Few researchers also proposed the novel parameters abstracted from the structure of natural genetics.

In 1995, D. B. Fogel had discussed the philosophy of mapping the parameters between the natural genetic structures and simulated environment of GA [26]. According to him, evolutionary computation can be conducted at various levels of abstraction such as genes and chromosomes in the GA. Simulated evolution can be made more biologically accurate by applying specific genetic operators that mimic low level transformations to DNA. This principle of abstraction from natural genetics has been attempted in the earlier work to design and develop novel primary twin operator for GA [27]. Later this primary twin operator was redesigned to introduce the advanced twin operator in order to improve the performance of GA.

In the earlier empirical study, the novel operator for GA called as advanced twin operator was proposed and developed in the simulated environment of GA [27] [28]. It was rigorously tested on the standard benchmark optimization test functions. In this way, the conventional GA is enriched with this novel advanced twin operator. This advanced twin operator design was originated by extrapolating the process of twin birth in natural genetics. This operator develops twin offspring individuals in the simulated environment of GA. This operator is associated with the design parameters viz. twin probability and twin separability. The twin probability resembles the frequency of producing twin offspring in natural reproduction systems and hence set to low value. The twin separability parameter is related to the number of unequal genes representing variation in twin's identical appearance and its phenotypic characteristics. In the earlier experimental work on this operator, the advanced twin operator was tested for static values of twin probability $p_{twin}$. The conclusion derived from this testing was the need to set and tune the parameter twin probability $p_{twin}$.

This paper introduces the novel adaptive approach of twin probability associated with the advanced twin operator that eliminates the burden on user of statistically defining the twin probability $p_{twin}$. The next section presents the design of the advanced twin operator and its parameters viz. twin probability and twin separability.

## II. DESIGN OF ADVANCED TWIN OPERATOR

The design of the advanced twin operator resembles the twin offspring formation due to single ovulation process in natural genetics. The process of generating twin individual strings involves the selection of fit parents for reproduction, crossover of these parents and then application of the advanced twin operator. The selection operator of GA is responsible for selecting the fit parents for reproduction.

Similar to natural system, one of the fit parents is equivalent to single ovum in the mating and reproduction process. The elitist strategy of GA requires that the best individual or the individual of first rank of the current generation should be forwarded as it is to the next generation. Therefore the second rank individual or individual next to the best individual should be selected for reproduction as the one of the parents resembling the single ovum in single ovulation. This parent is referred to as $p_1$. The other parent should be randomly selected from the current generation. This parent is referred to as $p_2$. The two parents are crossed using any conventional crossover method to generate two offspring say $child_1$ and $child_2$ respectively.

After this crossover, the advanced twin operator is designed and applied as follows. After reproduction, the hamming distance of each child from both the parents is calculated. Hamming distance indicate the unequal genes of the child from the respective parent. Consider that $H_1$ and $H_2$ are denoted as the hamming distances of $child_1$ from $p_1$ and $p_2$ respectively. The advanced twin operator should randomly select exactly half the number of unequal genes from $H_1$ as well as $H_2$ and change only the values of these genes by keeping all other genes same as that of $child_1$. This creates the first twin mate child say $child_3$. In this way, the advanced twin operator generates the first twin pair $child_1$:$child_3$.

The locations of unequal genes from $H_1$ and $H_2$ have considerable effect on the decoded value of the twin mate child. Similar process is repeated with $child_2$ to generate its twin mate $child_4$. This generates the second twin pair $child_2$:$child_4$. The design parameter of twin operation is the probability of twin operator say '$p_{twin}$'. It should be low as it resembles the frequency of twining in natural genetics. The other design parameter is number of unequal genes randomly selected from $H_1$ and $H_2$. It affects the twin separability parameter that increases in proportion to the number of unequal genes. The performance of GA varies in accordance with these design parameters.

## III. TWIN PROBABILITY AND ITS ADAPTIVE APPROACH

In the previous works, the advanced twin operator was tested with static values of $p_{twin}$ varying in the range of 0.02 to 0.1 on the various standard test functions [28]. In that case, it was necessary to set and tune twin probability in synchronization with other GA parameters viz. population size, selection operator, crossover probability $p_c$, mutation probability $p_m$ and the test problem to be solved. The drawbacks were the time needed to set and tune $p_{twin}$ and the burden on GA user to statistically define $p_{twin}$. To eliminate these drawbacks, this work proposes the adaptive approach of twin probability. The adaptive approach uses the fitness values of the best individual and the individual next to the best individual namely $f_{max}$ and $f_{max}'$.

It is proposed to change $p_{twin}$ in the following manner:
$$P_{twin} = f_{max} - f_{max}' \qquad (1)$$
provided that the following constraints are obeyed.
$$K_1 < f_{max} < K_1'$$
$$K_2 <= P_{twin} < K_3$$
where,

$K_1$ = 50% of the global optimum

$K_1'$ = 95% of the global optimum

$K_2$, $K_3$ = problem specific

Here, in the case of testing standard benchmark optimization functions, when the fitness values are scaled in the range 0 to 1, $K_2 = 0.05$ and $K_3 = 0.4$ are set.

## IV. EXPERIMRNTS AND RESULTS

The advanced twin operator and its adaptive approach of $p_{twin}$ is tested on the standard benchmark test functions as described and mentioned below:

### A. Himmelblau Function

This is a two variable unimodal function in the decoded parameter space often used as a test function in optimization literature [29]. In many engineering design problems, there is always need to find out the optimal set of design parameters that satisfy the number of goals at the same time. In these problems, each goal is formulated by the mathematical expression and the difference of expression from the target is calculated. The squares of the differences are taken and then summed up to form the objective function that is to be minimized. Therefore the Himmelblau function is the representative objective function of many engineering design problems.

The function is formally defined as:
$$F_1(x_1, x_2) = (x_1^2 + x_2 - 11)^2 + (x_1 + x_2^2 - 7)^2 \qquad (2)$$

with constraints: $0 <= x_1, x_2 <= 6$

In the feasible region, the function is unimodal in the decoded parameter space. The search space is considered in the range $0 <= x_1, x_2 <= 6$, in which this function has a single minimum point at (3, 2) with a function value equal to zero.

### B. Sphere Function

This function belongs to the standard De Jong's test suite. It is smooth, unimodal, symmetric and separable. It is the objective of every optimization algorithm to test its strength using this sphere function. The performance on this function is a measure of the general efficiency of the algorithm.

It is defined as:
$$F_2(X) = \sum_{i=1}^{n} x_i^2 \qquad (3)$$

with constraints: $-5.12 \leq x_i \leq +5.12$

This function has global minimum of zero for all values of $x_i = 0$, where $i = 1: n$.

### C. Rosenbrock Function

This function belongs to De Jong's test suite. It is a classic optimization problem, also known as Banana function. It has a very narrow ridge. The tip of the ridge is very sharp, and it runs around a parabola. The global optimum is inside a long, narrow, parabolic shaped flat valley. It is trivial to find the valley, however convergence to the global optimum is difficult and hence this problem has been repeatedly used in evaluating the performance of optimization algorithms. It is nonlinear, non-separable function.

It is defined as:
$$F_3(X) = \sum_{i=1}^{n-1} 100 \cdot (x_{i+1} - x_i^2)^2 + (1 - x_i)^2 \qquad (4)$$

with constraints: $-2.048 \leq x_i \leq +2.048$

This function has global minimum of zero for all values of $x_i = 1$, where $i = 1: n$.

### D. Rastrigin Function

This function is often used to test the genetic algorithm, because its many local minima make it difficult for standard, gradient-based methods to find the global minimum. This is based on the sphere function with the addition of cosine modulation to produce many local minima. Thus the function is highly multimodal. The locations of minima are regularly distributed.

It is defined as:
$$F_4(X) = 10.n + \sum_{i=1}^{n} (x_i^2 - 10 \cdot COS(2.\pi.x_i)) \qquad (5)$$

with constraints: $-5.12 \leq x_i \leq +5.12$

This function has global minimum of zero for all values of $x_i = 0$, where $i = 1: n$.

### E. Normalized Schwefel Function

The surface of Schwefel function is composed of a great number of peaks and valleys. The function has a second best minimum far from the global minimum where many search algorithms are trapped. Moreover, the global minimum is near the bounds of the domain. Schwefel's function is deceptive in that the global minimum is geometrically distant, over the parameter space, from the next best local minimum.

This function is multimodal and additively separable. Boundaries for this function are set at [-500,500].

It is defined as:
$$F_5(X) = \frac{\sum_{i=1}^{D} -x_i \sin(\sqrt{|x_i|})}{D} \qquad (6)$$

with constraints: $-5.12 \leq x_i \leq +5.12$, $\qquad i = 1,\dots,D$.

This function has global minimum at $x = 420.968$ with the function value equal to -418.9829.

In the design of experiments, Simple GA (SGA) and Advanced Twin GA (ATGA) with adaptive twin probability approach were designed and implemented for testing these optimization test functions. The selection of genetic operators and the setting of parameters for both SGA and ATGA were decided after carrying out lot of experimentation trials. The complete experimental setup for testing advanced twin operator on the standard benchmark optimization functions with the setting of important parameters is shown in Table I. For all functions to be tested, the setting and tuning of all the parameters of GA except twin probability was determined and kept same for SGA and ATGA. The stopping criterion was set to fixed number of generations. The details of the experimentation set up are described below for all the tested functions.

In the implementation of SGA designed for any function to be tested, the function variables were encoded in binary with the string length mentioned as per Table I and explained below. In the Himmelblau function implementation of GA, the two variables $x_1$ and $x_2$ were encoded in binary with the string of 40 bits. For Sphere function implementation, GA was designed with three variables $x_1$, $x_2$ and $x_3$ encoded in binary with the individual string length of 60 bits. In

Rosenbrock function, GA was designed with two variables encoded in binary with the individual string length of 40 bits.

In Rastringin multimodal function optimization, GA was designed with two variables encoded in binary giving the individual string length of 20 bits. In the testing of Normalized Schwefel function, GA was designed with two variables encoded in binary with string length of 44 bits. After extensive experimentation with its various settings SGA was designed with the population size of 40 and the reproduction operator of tournament selection with replacement. The single point crossover is used with the crossover probability $p_c$ set to 1. The fitness function was derived from the underlying objective function and then transformed to maximization. The termination criterion was set to fixed number of maximum generations as per Table I. SGA was executed several times with the different sets of static values of $p_m$ ranging from 0.001 to 0.05. After extensive experimentation, $p_m$ was set to 0.01 as it showed the best performance of SGA.

In the ATGA, the advanced twin operator was designed with adaptive approach of $p_{twin}$ as per the strategy mentioned in Equation 1. The twin separability parameter that indicates the unequal genes from hamming distances $H_1$ and $H_2$ was set to 50%. The results for both GAs were recorded by executing them for 25 repeated run trials in terms of performance parameters.

TABLE I.    EXPERIMENTAL SET UP FOR TESTING ADAPTIVE APPROACH

| Functions | Encoded String Length in bits | Pop_size | Selection | Crossover ($p_c$=1) | Fixed $Pm$ | Max. gens |
|-----------|-------------------------------|----------|-----------|---------------------|------------|-----------|
| Himmelblau | 20 | 40 | Tournament with replacement | Single point | 0.01 | 15 |
| Sphere | 60 | 40 | Tournament with replacement | Single point | 0.01 | 15 |
| Rastringin | 20 | 40 | Tournament with replacement | Single point | 0.01 | 15 |
| Rosenbrock | 40 | 40 | Tournament with replacement | Single point | 0.01 | 15 |
| Normalized Schwefel | 44 | 40 | Tournament with replacement | Single point | 0.01 | 20 |

TABLE II.    SUMMARISED RESULTS

| Parameters | Himmelblau | | Sphere | | Rastringin | | Rosenbrock | | Normalized Schwefel | |
|-----------|------------|------|--------|-------|------------|-------|------------|-------|---------------------|--------|
| | *SGA* | *ATGA* | *SGA* | *ATGA* | *SGA* | *ATGA* | *SGA* | *ATGA* | *SGA* | *ATGA* |
| Mean of Best Individual | 0.72 | 0.80 | 0.81 | 0.84 | 0.41 | 0.49 | 0.80 | 0.83 | 394.46 | 405.11 |
| Max. of Best Individual | 0.98 | 0.99 | 0.98 | 0.999 | 0.99 | 0.99 | 0.998 | 0.999 | 414.21 | 418.63 |
| Coefficient of Variance | 31 | 24 | 17 | 14 | 36 | 34 | 17 | 15 | 6 | 5 |
| Mean Convergence Generation | 9 | 8 | 10 | 8 | 8 | 8 | 10 | 8 | 14 | 12 |

The important performance measures recorded were the best individual, mean values of the best individual, generation wise average fitness and the average value of convergence generation. The basic statistical analysis of the best individual was carried out in terms of mean value, maximum value and coefficient of variance. This basic statistical analysis along with the mean convergence generation for all the tested functions is displayed in Table II. The generation wise average fitness is recorded for all test functions. The variations of the average fitness per generation are displayed in Fig. 1, Fig. 2, Fig. 3, Fig. 4 for the functions of Himmelblau, Sphere, Rastringin and Rosenbrock respectively. Besides this, the best individuals were recorded for 25 repeated run trials for all the tested benchmark functions. The graphs of the best individual recorded for 25 repeated run trials in SGA and ATGA are displayed in Fig. 5 for the Sphere function and in Fig. 6 for Rastringin function respectively. Similar results were also observed for the rest of the tested functions.

## V.  CONCLUSION AND FUTURE SCOPE

It is observed from Table II that the mean value of the best individual with ATGA is better than that of SGA. Besides, the coefficient of variance due to ATGA is lower than that of SGA. This indicates the improvement in the performance of GA in terms of the best individual. There is no significant comparative change in the maximum value of the best individual resulted due to both GAs except Normalized Schwefel function. The results for this function are displayed as absolute values of the optimum in Table II. These results clearly indicate better values of both the mean best individual and maximum best individual for ATGA than that of SGA.

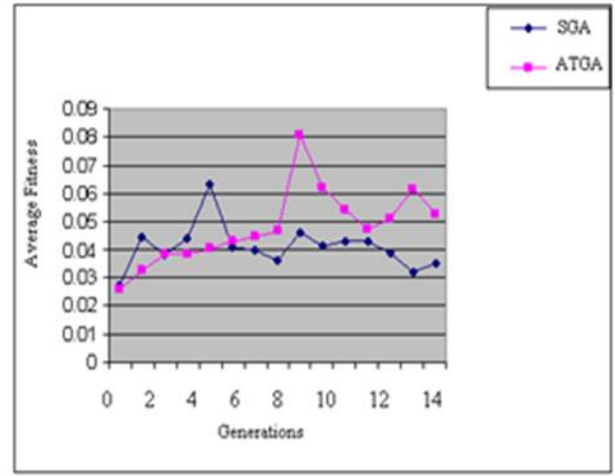

Figure 3.  Generation wise Average Fiteness for Rastringin Function

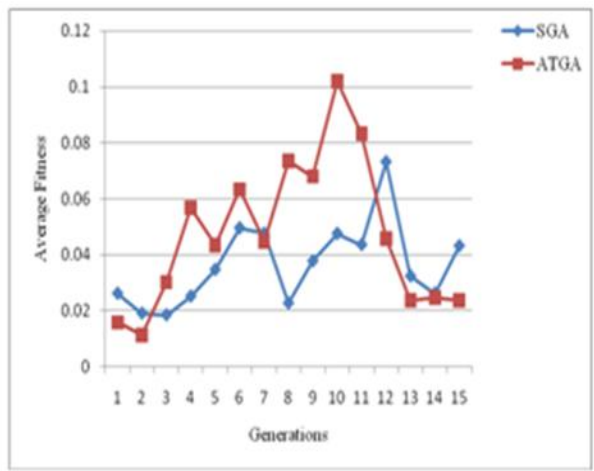

Figure 1.  Generation wise Average Fiteness for Himmelblau Function

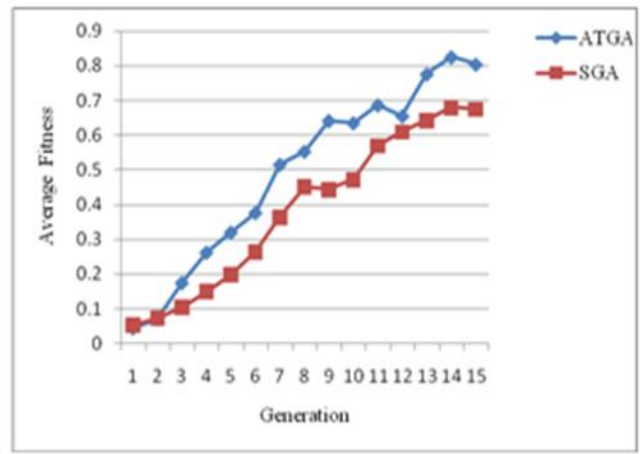

Figure 4.  Generation wise Average Fiteness for Rastringin Function

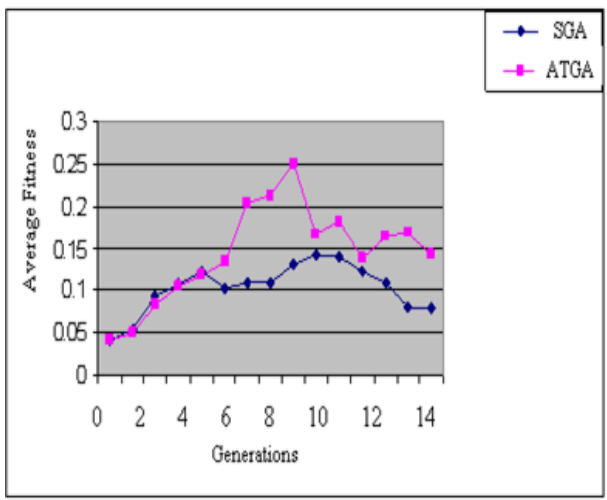

Figure 2.  Generation wise Average Fiteness for Sphere Function.

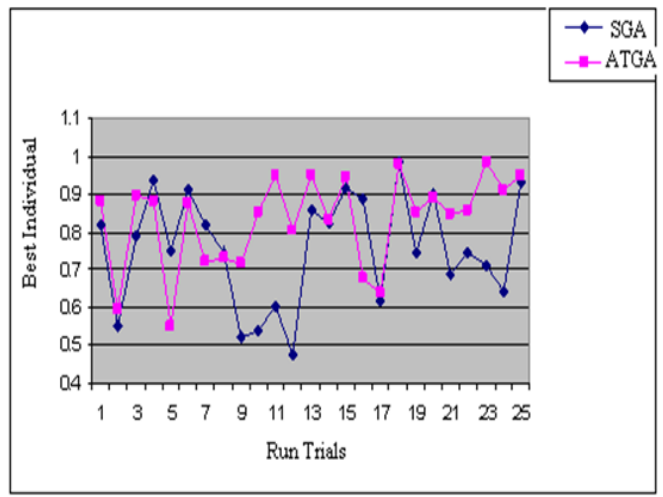

Figure 5.  Generation wise Average Fiteness for Rastringin Function

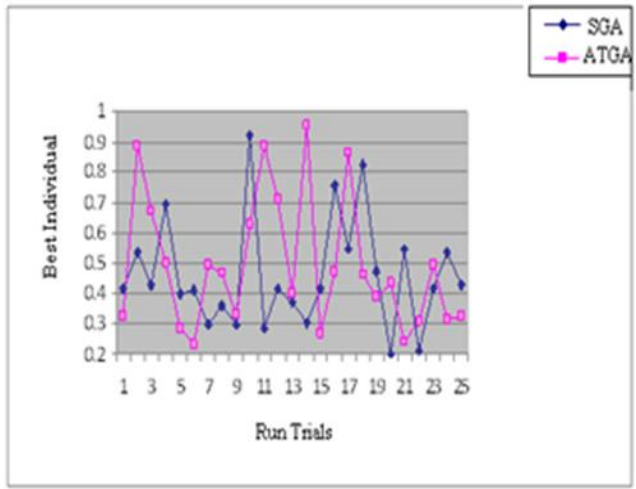

Figure 6.   Best Individual Vs GA Run Trials for Sphere Function

In addition to this, the mean convergence generation required by ATGA is slightly lower than that of SGA for all test functions except Rastringin indicating the comparatively reduced functional computations because of the advanced twin operator with adaptive twin probability approach. The graphs displayed in Fig. 1, Fig. 2, Fig. 3, Fig. 4 highlight the rise in average fitness of ATGA showing the effect of the advanced twin operator. Similarly it is also observed that the best individual values due to ATGA are better than that of SGA for most of the repeated run trials as displayed in Fig. 5 and Fig. 6 respectively. After extensive experimentation and testing, it may be concluded that this adaptive twin probability approach of the advanced twin operator improves the overall performance of GA and also eliminates the time consuming operation of statically setting and tuning the twin probability.

The current research work may be extended in future by attempting the novel adaptive approaches for twin probability based on the concept of classifying the various GA parameters. The classifiers may be designed on the basis of few important GA parameters viz. population distribution, average fitness, genotype structure. The design approach of classifiers may be based on the defined rule based systems or neural network. Further the hybrid methods can also be developed for designing high performance classifiers. In addition to this, the current work may also be extended by incorporating the adaptive approaches for the probabilities of crossover and/or mutation. Further it will be necessary to tune these approaches with the adaptive approach of twin probability. This task of tuning will be highly critical as there exists the number of adaptive approaches for tuning crossover and mutation probabilities in GA literature.

From the analysis of results and discussions, it is summarized that the effectiveness of this adaptive twin probability may be verified in real world applications to solve the complex and difficult search and optimization problems in engineering and industry.

### REFERENCES

[1]   Goldberg D. E., "Genetic Algorithms in Search, Optimization and MachineLearning", Reading MA: Addison Wesley, 1989.

[2]   Pinaki Mazumder, Elizabeth M. Rudnik, Genetic Algorithm for VLSI Design, Layout and Automation, Pearson Education (Singapore), Delhi, India, 1999.

[3]   Rajasekaran S., Vijayalakshmi Pai G. A., Neural Networks, Fuzzy Logic, and Genetic Algorithms: Synthesis and Applications, Prentice Hall of India, New Delhi, 2003.

[4]   Earl Cox, Fuzzy Modeling and Genetic Algorithms for Data Mining and Explorator, Elsevier, 2005.

[5]   Spears W.M, "A Study of Crossover Operators in Genetic Programming, in International Symposium on Methodologies for Intelligent Systems," pp. 409-418, 1991.

[6]   Andrew Czarn, Cara MacNish, Kaipillil Vijayan, Berwin Turlach, and Ritu Gupta, "Statistical Exploratory Analysis of Genetic Algorithms", IEEE Transactions Evolutionary Computation, vol. 8, No. 4, pp. 405-421, August 2004.

[7]   Esquivel S., Gallard R. and Michalewicz Z., "Another Approach to crossover in Genetic Algorithms", Proc. of Primer Congreso Argentino De Ciencias De La Computation, pp. 141-150, 1995.

[8]   Esquivel S., Leiva A., Gallard R., "Multiple Crossover per Couple in Genetic Algorithms", Proc. 4th IEEE conference on Evolutionary computation, Indianapolis, USA, 1997.

[9]   Esquivel S., Leiva H., Gallard, R., "Self-Adaptation of Parameters for MCPC in Genetic Algorithms", Proc. 4th Congreso Argentino de Ciencias de la Computación , pp. 419-425, Universidad Nacional del Comahue, 1998.

[10]  Esquivel S., Leiva A., Gallard R., "Multiple Crossovers between Multiple Parents to improve search in Evolutionary Algorithms", Proc. Congress on Evolutionary Computation, vol.2, pp. 1594, 1999.

[11]  Gallard R. H., Esquivel S. C., "Enhancing Evolutionary Algorithms through Recombination and Parallelism", J. Computer Science and Technology, 2001

[12]  Eiben A.E., Raué P-E., Ruttkay Zs., "Genetic Algorithms with Multiparent Recombination", Proc. 3rd Conference on Parallel Problem Solving from Nature, pp.78-87, 1994.

[13]  F. Herrera, M. Lozano, E. Pérez, A.M. Sánchez, P. Villar, "Multiple Crossover per Couple with Selection of the Two Best Offspring: An Experimental Study with the BLX-α Crossover Operator for Real-Coded Genetic Algorithms", Proc. Ibero-American Conference on Advances in Artificial Intelligence, Sevilla (Spain), pp 392-401, 2002.

[14]  Carballido, J., Ponzoni, I., Brignole, N. B., "MCPC and MCMP Evolutionary Algorithms for the TSP", Proc. Argentine Symposium on Artificial Intelligence,2003

[15]  De San Pedro, M.E. Pandolfi, D. Villagra, A. Lasso, M. Gallard, R.H. "Effect of crossover operators under Multirecombination: Weighted Tardiness, a Test Case", Proc. Congress on Evolutionary Computation, vol. 1, pp. 699-705,2004.

[16]  Vladimir Estivill-Castro, "Adaptive Genetic Operators", IEEE proceedings of Intelligent Information Systems, IIS'97, pp. 194-198, 8-10 Dec. 1997.

[17]  Yoon H. S., Moon B. R., "An Empirical Study on the Synergy of Multiple Crossover Operators", IEEE Transactions on Evolutionary Computation, 6(2), pp. 212-223, 2002.

[18] Davis L., "Adapting operator Probabilities in genetic algorithms," Proceedings of the Third International Conference on Genetic Algorithms, 1989, pp. 61-69.

[19] Srinivas M. and Patnaik L., "Adaptive Probabilities of Crossover and Mutation in Genetic Algorithms", IEEE Transactions on Syst, Man and Cybernetics, vol. 24, pp. 656-667, 1994.

[20] Jun Zhang, H.S.H. Chung, B. J. Hu., "Adaptive Probabilities of Crossover and Mutation in Genetic Algorithms based on Clustering Technique", Congress on Evolutionary Computation, Portland, vol. 2, pp. 2280-2287, 2004.

[21] Dai C.H., Zhu Y.F., Chen W.R., "Adaptive Probabilities of Crossover and Mutation in Genetic Algorithms based on Cloud Model", IEEE Information Theory Workshop, pp. 710-713, 2006.

[22] Jun Zhang, Henry Shu-Hung Chung and Wai-Lun Lo, "Clustering Based Adaptive Crossover and Mutation Probabilities for Genetic Algorithms", IEEE Transactions on Evolutionary Computation, vol. 11, pp. 3, 2007.

[23] Fogarty T. C., "Varying the Probability of Mutation in Genetic Algorithms," Proceedings of Third International Conference on Genetic Algorithms, pp. 104-109, 1989.

[24] Cervantes J. and Stephens C. R., "Limitations of Existing Mutation Rate Heuristics and How a Rank GA Overcomes Them", IEEE Transactions on Evolutionary Computation, 13 (2), April 2009.

[25] Matej Crepinsek, Shih-His Liu, Marjan Mernik, "Exploration and Exploitation in Evolutionary Algorithms: A Survey", ACM Computing Surveys, Vol. 1, No. 1, Article A, January 2011.

[26] David B. Fogel, 1995, "Phenotypes, Genotypes and Operators in Evolutionary Computation", Proc. 2nd IEEE International Conference on Evolutionary Computation, pp. 193-198.

[27] Anagha Parag Khedkar, Subbaraman Shaila, 2010, "Novel Twin Operator for Genetic Algorithm", Proc. of 3rd International conference on Data Management: Innovations and Advances in Computer Science and Engineering, pp. 111-117.

[28] Anagha Parag Khedkar, Shaila Subbaraman, "Effect of Advanced Twin Operator on the Performance of Genetic Algorithm", International Journal of Engineering Research and Technology, India, Vol. 3, pp. 721-731, 2010.

[29] Deb K., Optimization for Engineering Design: Algorithms and Examples, Prentice Hall: New Delhi, 1995.